\documentclass[11pt]{amsart}
 
\usepackage{graphicx,algorithm,algorithmic,amsmath,amssymb}
\RequirePackage[dvips]{hyperref}
\usepackage[sort]{natbib}
\RequirePackage{hypernat}
\bibpunct{(}{)}{;}{a}{,}{,}

\usepackage[left=3cm,right=3cm,top=3cm,bottom=3cm]{geometry}

\def\b{\beta}
\def\l{\lambda}
\def\a{\alpha}
\def\R{\mathbb{R}}
\def\s{\mathbf{s}} 
\def\sgn{\mathrm{sign}}
\def\sign{\sgn}
\def\diag{\mathrm{diag}}

\def\d{\delta}

\def\posprop{\stackrel{+}{\propto}}

\newtheorem{theorem}{Theorem}
\newtheorem{corollary}{Corollary}
\newtheorem{example}{Example}
\newtheorem{remark}{Remark}

\title{Exact block-wise optimization in group lasso and sparse group lasso for linear regression}
\author{Rina Foygel \and Mathias Drton}
\address{Department of Statistics, The University of Chicago, Chicago,
  IL, U.S.A.}
\email{rina@uchicago.edu, drton@uchicago.edu}

\begin{document}

\begin{abstract}%
The group lasso is a penalized regression method, used in regression problems where the covariates are partitioned into groups to promote sparsity at the group level. Existing methods for finding the group lasso estimator either use gradient projection methods to update the entire coefficient vector simultaneously at each step, or update one group of coefficients at a time using an inexact line search to approximate the optimal value for the group of coefficients when all other groups' coefficients are fixed. We present a method of computation for the group lasso in the linear regression case, the Single Line Search (SLS) algorithm, which operates by computing the exact optimal value for each group (when all other coefficients are fixed) with one univariate line search. We perform simulations demonstrating that the SLS algorithm is often more efficient than existing computational methods. We also extend the SLS algorithm to the sparse group lasso problem via the Signed Single Line Search (SSLS) algorithm, and give theoretical results to support both algorithms.
\end{abstract}


\maketitle

\section{Introduction}

Consider a normal regression problem with response vector $y\in\R^n$ and covariate matrix $X\in\R^{n\times p}$ that is decomposed into `blocks' or `groups', as $X=(X_1 \ X_2 \ \dots \ X_K)$. In this linear regression setting, the group lasso estimator \citep{yuan2006model,kim2006blockwise} is a coefficient vector $\b\in\R^p$ which minimizes the objective function
\begin{equation}\label{eqn:GL}L(\b)=\tfrac{1}{2}\|y-X\b\|_2^2+\l\sum_{k=1}^K\|\b_k\|_2\;\;,\end{equation}
where $\l>0$ is a penalty parameter. This penalized regression method addresses the problem of model selection when the true model is believed to be `groupwise sparse'; that is, when the smallest true model might plausibly exclude some of the groups $\{X_1,\dots,X_K\}$ entirely. This setting is found in many applications, since covariates are often naturally grouped in some manner. Each group might contain a number of factor levels of a single factor (for instance, in genetic data, the factor could indicate the presence of zero, one, or two copies of a rare allele), or might consist of a set of related quantitative variables. It has been shown that the group lasso objective function can be applied to accurately and efficiently recover group-wise sparse signals \citep{yuan2006model,meier2008group,kim2006blockwise}, and that the group lasso estimator shows asymptotic consistency even when model complexity grows with sample size \citep{nardi2008asymptotic}. The group lasso has been discussed in many settings aside from normal regression, including logistic regression \citep{meier2008group} and generalized linear models \citep{kim2006blockwise}. The group lasso has been applied to multivariate regressions, where the response variables are expected to have similar or identical sparsity patterns and therefore the matrix of coefficients is likely to be row-wise sparse, with asymptotic consistency results \citep{obozinski2008high}. Similar objective functions have been proposed to handle a range of settings, including the possibility of overlapping groups \citep{jacob2009group}, and multi-task learning \citep{obozinski2010joint,lounici2009taking}.

We remark that it is common to include an unpenalized intercept term or other unpenalized terms in the group lasso. However, such an objective function can be reduced to the form of~(\ref{eqn:GL}) by regressing out all unpenalized covariates from the response and the penalized covariates. Additionally, there are many settings where we might wish to place different positive penalties on the different groups, giving the more general objective function:
$$L(\b)=\tfrac{1}{2}\|y-X\b\|_2^2+\sum_{k=1}^K\l_{(k)}\|\b_k\|_2\;\;.$$
For instance, penalties are scaled with the square root of group size to penalize larger groups in~\cite{meier2008group}. However, rescaling the groups of covariates can transform this objective function into the form of~(\ref{eqn:GL}). Therefore, in this paper we only consider the case where each group has equal positive penalty $\l$.

Recently, the sparse group lasso was proposed as an extension to the group lasso, placing an additional penalty on the $1$-norm of the coefficient vector \citep{wu2008coordinate,friedman2010note}. The objective function is then given by 
\begin{equation}\label{eqn:SGL}L_1(\b)=\tfrac{1}{2}\|y-X\b\|_2^2+\l_1\sum_{k=1}^K\|\b_k\|_2+\l_2\|\b\|_1\;\;.\end{equation}
As in the group lasso problem~(\ref{eqn:GL}) we assume that the penalty $\l_1$ on group norms is positive; we may also assume $\l_2>0$, since~(\ref{eqn:SGL}) reduces to~(\ref{eqn:GL}) if $\l_2=0$. \cite{friedman2010note} argue that this method is more appropriate when there is the possibility of within-group sparsity, and show that optimizing the objective function does in fact recover both group-wise and within-group sparsity in simulations. As with the group lasso, a sparse group lasso problem with unpenalized covariates may be reduced to an objective function in the form of~(\ref{eqn:SGL}).

The solution to a group lasso or sparse group lasso problem is not necessarily unique. As a simple example, consider the case of repeated groups, where $X_{k_1}=X_{k_2}U$ for two groups $k_1\neq k_2$ and some orthogonal matrix $U$. This may produce an infinite solution set. (Or, if the penalties vary across the groups, we might have $\l_{(k_1)}^{-1}X_{k_1}=\l_{(k_2)}^{-1}X_{k_2}U$ for some orthogonal $U$). However, the minimum of the objective function is always attained, and there is a unique optimal vector of fitted values (denoted by $\hat{y}$ in this paper) and a unique penalty term value. In other words, in the group lasso case \citep{roth2008group},
$$\hat{\b}^1,\hat{\b}^2\in\hat{B} \ \Rightarrow \ X\hat{\b}^1=X\hat{\b}^2 \ \mathrm{and} \ \sum_{k=1}^K\|\hat{\b}^1_k\|_2=\sum_{k=1}^K\|\hat{\b}^2_k\|_2\;\;,$$
where $\hat{B}=\arg\min_{\b}L(\b)$. By analagous reasoning, the same is true for the sparse group lasso case with objective function $L_1$. Furthermore we can say that the `direction' of $\hat{\b}_k$ for each group $k$ is unique. The precise meaning of this is explained in the following theorem:

\begin{theorem}\label{thm:solution_set_GL}
Let $\hat{B}$ be the set of minimizers of the penalized likelihood $L$ for a group lasso problem~(\ref{eqn:GL}) or sparse group lasso problem~(\ref{eqn:SGL}). Then there exists a unique minimal set of groups $\mathcal{K}\subset\{1,\dots,K\}$, and unique unit vectors $v_k\in\R^{p_k}$ for each $k\in\mathcal{K}$, such that
$$\hat{\b}\in\hat{B} \ \Rightarrow \ \hat{\b}_k\posprop v_k \ \forall k\in\mathcal{K}, \ \hat{\b}_k=0 \ \forall k\not\in\mathcal{K}\;\;,$$
where we define $a\posprop b$ to mean that $a=c\cdot b$ for some nonnegative scalar $c$.
\end{theorem}

Many advances have been made in recent years for efficient optimization of the group lasso penalized likelihood function. The algorithms may be broken down into two broad categories: group-wise descent, where each step updates one entire group of coefficients via an inexact line search \citep{meier2008group}, and `global' descent, where at each step the entire coefficient vector could potentially be updated \citep{kim2006blockwise,roth2008group}. An efficient approach to the corresponding online learning problem, developed by~\cite{yang-online}, handles the online versions of both the group lasso and sparse group lasso. Each step of the online learning algorithm is very efficient, and the algorithm requires no precomputations. Since online learning is a very different task from the offline convex optimization problem which we seek to solve, we do not attempt to compare this algorithm to the others in our simulations.

The main result of this paper is the `Single Line Search' (SLS) algorithm for solving the group lasso problem. The efficiency of this method lies in the computation of the exact optimal value for the coefficients of any single group (fixing the other coefficients) via a single univariate line search, which corresponds to finding the $2$-norm of the optimal coefficient vector for that group. We state several theoretical results, showing that each group's update is indeed optimal, that the algorithm converges to the minimum of the objective function, and that at any finite time in the algorithm the distance to convergence can be bounded in terms of the present subgradient norm. We present simulation results showing that this method performs faster on the group lasso problem than existing offline learning algorithms (including both group-wise and global descent algorithms).

Turning to the sparse group lasso problem, we extend the SLS algorithm to handle the additional $\ell_1$ penalty on the coefficient vector. This method is, to our knowledge, the only existing algorithm for solving the sparse group lasso problem in its `offline' form. The structure of the SSLS algorithm makes it practical only when the groups' sizes are quite small; therefore, we discuss strategies for developing an efficient algorithm to solve the sparse group lasso problem which is more flexible with respect to group size. We also discuss possible extensions to the SLS algorithm, including extending the algorithm to models other than linear regression, and adaptations to the algorithm which may increase efficiency in extremely high-dimensional group lasso settings (as examined in~\citealp{roth2008group}, for instance).

The remainder of the paper is structured as follows. In Section~\ref{section:PriorWork}, we summarize existing methods for computing the group lasso solution. In Section~\ref{section:SLS_theory}, we introduce the SLS algorithm and the main related theoretical results. Results of simulations comparing the SLS algorithm to existing methods are given in Section~\ref{section:SLS_results}. In Section~\ref{section:SSLS}, we introduce the SSLS algorithm for the sparse group lasso and give theoretical results, and also describe the existing algorithm for solving the online version of the problem. Section~\ref{section:Discussion} contains the discussion of our results and of future directions. Unless otherwise noted, all theoretical results in the paper, including Theorem~\ref{thm:solution_set_GL} above, are proved in the Appendix (Section~\ref{section:Appendix}).

Since making this manuscript publicly available, we have been made
aware of the earlier work by~\citet{puig2009multidimensional}, which derives the same
result for the (non-sparse) group lasso setting. We leave this manuscript available
as a technical report, to serve as a reference for the previously
untreated sparse group lasso case (the SSLS algorithm), and for the
timing comparisons of various methods in the group lasso setting.

\section{Prior work}\label{section:PriorWork}

In this section we outline prior work on the group lasso problem, consisting of both group-wise descent and global descent methods.

\subsection{Group-wise descent}

We first examine existing computations and methods for group-wise descent. Since only one group at a time is being updated, we may restrict our attention to the subproblem of finding $\b_k$ to minimize
$$\tfrac{1}{2}\|R_k-X_k\b_k\|^2_2+\lambda\|\b_k\|_2\;\;,$$
where $R_k$ is the remainder when all other coefficients are fixed, $R_k=y-\sum_{l\neq k}X_l\beta_l$. For simplicity of notation, we change variables and write this subproblem as the minimization of
\begin{equation}\label{eqn:subproblem_GL}Q(\a)=\tfrac{1}{2}\|b-A\a\|_2^2+\l\|\a\|_2\;\;,\end{equation}
where $b\in\R^n$, $A\in\R^{n\times q}$, $\l>0$.

The objective function~(\ref{eqn:subproblem_GL}) is clearly convex in $\a$, and by Theorem~\ref{thm:solution_set_GL} has a unique minimizer $\hat{\a}$. A subgradient of $Q$ at $\a$ is any vector
$$-A^Tb+A^TA\a+\l s\;\;,$$
where $s=\a/\|\a\|_2$ for nonzero $\a$, or may be any vector of up to unit norm when $\a=0$ \citep{bertsekas1999nonlinear,meier2008group}. The subdifferential of $Q$ at $\a$, written $\partial Q(\a)$, is the set of all subgradients at $\a$. Since $Q$ is convex, the subgradient condition for optimality shows that $\a$ is optimal if and only if $0\in\partial Q(\a)$.

It is clear from the known subgradient condition that
$$0\in\arg\min_{\a}Q(\a) \ \Leftrightarrow \ \|A^Tb\|_2\leq \l\;\;.$$
The zero case is therefore simple and we turn to the case that $\a=0$ is not optimal.

\cite{yuan2006model} give the solution to the subproblem in the case where the columns of $A$ are orthonormal. In this case, examining the subdifferential shows that $\a\neq 0$ is optimal if and only if
$$0=-A^Tb+\a+\l \frac{\a}{\|\a\|_2} \ \Leftrightarrow \ \left(1+\frac{\l}{\|\a\|_2}\right)\a=A^Tb \ \Leftrightarrow \ \a=\left(1-\frac{\l}{\|A^Tb\|_2} \right)A^Tb\;\;.$$
Computing this last quantity is very fast, and we may use it in any setting to compute a group-wise sparse regression by orthonormalizing each group of covariates $X_k$. However, \cite{friedman2010note} raise the point that the resulting solution, when transformed back to the original basis for each group, will not be a solution of the group lasso problem with the original covariates. In many situations, orthonormalizing each group of covariates may be unnatural or undesirable. Therefore, methods which do not require orthonormal $X_k$'s are necessary.

For the general case, where covariate groups are not assumed to be orthonormalized, an iterative procedure updating coefficients one group at a time is proposed by \cite{meier2008group}, and implemented in the R package \verb#grplasso# for both the linear and logistic regression settings. A rough sketch of their method is as follows. Each iteration of the algorithm cycles through the groups, updating $\b_k$ via a quasi-Newton method. Specifically, at group $k$, holding the coefficients for all other groups fixed, the algorithm will seek to improve the estimate of $\b_k$ as follows. Let $\b^0$ denote the present value of the coefficient vector. First, the (unpenalized) negative likelihood function is approximated, near $\b^0_k$, via a quadratic function in $d=\b_k-\b^0_k$, with quadratic term $c\cdot d^Td$ for some scalar $c>0$. (Note that, in the linear regression setting, the negative likelihood function is always a quadratic form with positive semidefinite leading coefficient matrix; however, the coefficient matrix in the quadratic term might not be of the form $cI_{p_k}$ for any $c$.) The objective function is then approximated by adding the penalty term. Finally, the algorithm computes a minimizer $\hat{d}$ of the approximated objective function, and updates $\b_k=\b^0_k+s\cdot \hat{d}$, where the scalar $s$ is chosen via an (inexact) Armijo line search (or updates $\b_k=0$ if appropriate).

This method's effectiveness lies in the efficiency of the updates, and in the fact that when the current estimate is quite close to the optimum, the update to each group of coefficients is a very good approximation of the true optimum for that group of coefficients when the other groups are fixed. The algorithm also requires very little precomputation.

\subsection{Global descent}
Next, we examine existing methods which update the entire coefficient vector simultaneously at each step. \cite{kim2006blockwise} propose one such method, in which the sum of the groups' $2$-norms is bounded rather than penalized:
\begin{equation}\label{eqn:GL_bounded}\arg\min\left\{\|y-X\b\|^2_2 \ : \ \sum_{k=1}^K\|\b_k\|_2\leq M\right\}\;\;.\end{equation}
(This is equivalent to placing a penalty of $\l$ on $\sum_{k=1}^K\|\b_k\|_2$ for some $\l\geq0$ whose exact value depends both on $M$ and on the data). In the terminology of Section 2.3 of~\cite{bertsekas1999nonlinear}, \citeauthor{kim2006blockwise}'s \citeyearpar{kim2006blockwise} algorithm is a gradient projection method with constant stepsize. A brief outline of the algorithm is as follows. Given an estimate of $\b$, the algorithm first computes the gradient of the (unpenalized) likelihood function, and takes a small step along that gradient. Next, this resulting vector is projected to the closest vector satisfying the bound condition on the sum of group norms. This process is then repeated until convergence.

\cite{roth2008group} propose a modification of \citeauthor{kim2006blockwise}'s~\citeyearpar{kim2006blockwise} algorithm, which makes use of group-wise sparsity for faster convergence. At each iteration, the algorithm has some hypothesized `active set' of groups which are currently included in the model. The coefficient vector is then optimized over that active set alone, using \citeauthor{kim2006blockwise}'s \citeyearpar{kim2006blockwise} optimization algorithm. Once convergence on the active set is reached, the solution is tested for optimality; if it fails, then the active set is updated based on that information, and the procedure is repeated. This algorithm may be particularly efficient when there is an optimal solution involving only a very small fraction of the groups of covariates. In particular, their experiments show improved time by several orders of magnitude in such scenarios.

Overall, a global search algorithm may be particularly efficient when there is high correlation between the groups of coefficients, because group-wise descent may result in `zig-zag' paths to the optimum in this type of setting.

\section{The SLS algorithm for the group lasso}\label{section:SLS_theory}

We first state a result which motivates our method.

\begin{theorem}
\label{thm:GL}
Define
$$Q(\a)=\tfrac{1}{2}\|b-A\a\|_2^2+\l\|\a\|_2\;\;,$$
where $b\in\R^n$, $A\in\R^{n\times q}$, $\l>0$, and $\a$ may take any value in $\R^q$.
Let $A^TA=U^TDU$ be the spectral decomposition, with $D=\diag\{d_1,d_2,\dots,d_q\}$. Define $v=UA^Tb$. Then:
\begin{itemize}
\item[i.] If $\|v\|_2\leq \lambda$ then $\alpha=0$ is the unique minimizer of $Q$.
\item[ii.] If $\|v\|_2>\lambda$, then there is a unique $r\in\mathbb{R}_+$ satisfying
$$f(r)=\sum_{j=1}^{q} \frac{v_j^2}{(d_jr+\lambda)^2}=1\;\;.$$
Furthermore, if we define
$$\alpha(r)=U^T(D+r^{-1}\lambda I_p)^{-1}v\;\;,$$
then $\alpha=\alpha(r)$ is the unique minimizer of $Q$.
\end{itemize}
\end{theorem}

We are now ready to define the Single Line Search (SLS) algorithm; see the pseudocode in Algorithm~\ref{alg:SLS}. The intuition for the algorithm is simple. During each iteration, we cycle once through the groups. At group $k$, we fix the coefficients outside of the group and compute the partial residual $R_k=y-\sum_{l\neq k}X_l\b_l$. We then apply Theorem~\ref{thm:GL} to find the exact optimal value for $\b_k$, given the fixed coefficients outside the group. (Specifically, in the notation of Theorem~\ref{thm:GL}, we may easily solve for $r$ using Newton's method, since $f(r)$ is a strictly decreasing function with a derivative that is simple to compute). This strategy involves more pre-computation than the existing algorithms, as it requires a spectral decomposition for any group which is included in the model at any stage of the algorithm. In the scenarios we consider in our simulations, however, this one-time computational cost is outweighed by the efficiency of each group's update.

An immediate corollary to Theorem~\ref{thm:GL} is the following:
\begin{corollary}\label{cor:decrease_GL}
Let $\b^{(t)}$ be the coefficient estimate after $t$ iterations of the SLS algorithm for $t=0,1,2,\dots$. Then for all $t$,
$$L(\b^{(t+1)})\leq L(\b^{(t)})\;\;.$$
That is, each iteration of the algorithm does not increase the objective function.
\end{corollary}

Next we state a convergence result for this algorithm, which follows directly from Proposition 5.1 of ~\cite{tseng2001convergence}. Note that no conditions are necessary on the data $(X,y)$ or the (positive) penalty $\l$.

\begin{theorem}\label{thm:convergence_GL}
Let  $\b^{(t)}$ be the coefficient vector after the $t^{th}$ iteration of Algorithm~\ref{alg:SLS}. Then $L(\b^{(t)})\rightarrow\min_{\b}L(\b)$.
\end{theorem}

Finally, since in practice we will wish to terminate the algorithm after a finite number of steps, the following theorem gives a guarantee of accuracy. When terminating the algorithm after $t$ iterations, we can apply the theorem below (with $\b^*=\b^{(t)}$) to bound the error in the current estimate of the optimal fitted values $\hat{y}$.

\begin{theorem}\label{thm:accuracy_GL}
Take any $\b^*\in\R^p$, and any $w^*\in \partial L(\b^*)$. Let $\hat{y}$ be the unique optimal vector of fitted values. Then
$$\|X\b^*-\hat{y}\|^2_2\leq 2(w^*)^T\b^*+\mathbf{O}(\|w^*\|_2)\;\;,$$
with precise bounds given in the Appendix.
\end{theorem}

\begin{algorithm}[t]
\caption{Single Line Search (SLS) algorithm for the group lasso}
\label{alg:SLS}
\begin{algorithmic}
\STATE \textbf{Input:} $y\in\R^n$, $X_1\in\R^{n\times p_1},\dots, X_K\in\R^{n\times p_K}$, $\l>0$.
\STATE \textbf{Output:} $\b\in\R^p$ minimizing $L(\b)=\tfrac{1}{2}\|y-X\b\|_2^2+\l\sum_{k=1}^K\|\b_k\|_2$, where $p=p_1+\dots+p_K$ and $X=(X_1 \ \dots \ X_K)$.\\
\STATE \textbf{Initialize:} $\b\Leftarrow \b^{(0)}$.\\
\REPEAT
\FOR{$k=1,2,\dots,K$} 
\STATE $R_k\Leftarrow y-\sum_{l\neq k}X_l\b_l$.
\IF{$\|X_k^TR_k\|_2\leq \l$}
\STATE $\b_k\Leftarrow 0$.
\ELSE
\STATE Compute the spectral decomposition $X_k^TX_k=U_k^TD_kU_k$ if not previously computed, and write $D_k=\diag\{d^k_1,\dots,d^k_{p_k}\}$.
\STATE $v_k\Leftarrow U_kX_k^TR_k$.
\STATE Find the unique $r>0$ satisfying $f(r)=\sum_{j=1}^{p_k}\frac{(v_k)_j^2}{(d^k_jr+\l)^2}=1$.
\STATE $\b_k\Leftarrow U_k^T(D_k+r^{-1}\l I_{p_k})^{-1}v_k$.
\ENDIF
\ENDFOR
\UNTIL{some convergence criterion is met.}
\end{algorithmic}
\end{algorithm}

\section{Simulations for the group lasso}
\label{section:SLS_results}
We compare the speed of the SLS algorithm to the three existing methods described above: \citeauthor{meier2008group}'s \citeyearpar{meier2008group} group-wise search algorithm, \citeauthor{kim2006blockwise}'s \citeyearpar{kim2006blockwise} bounded global-update algorithm, and \citeauthor{roth2008group}'s \citeyearpar{roth2008group} active-set modification of \citeauthor{kim2006blockwise}'s \citeyearpar{kim2006blockwise} method.

Our simulations vary along three different parameters: the total number of groups, $K$; the level of within-group correlation, $a$; and the level of between-group similarity, $b$ (each described in detail below). For each parametrization, we run 100 trials. In each trial, we generate covariates $X$ and response $y$, and also a decreasing sequence of 5 penalty parameters $\{\l^1,\dots,\l^5\}$. We run each of the four algorithms on sequence of group lasso problems defined by the data $(X,y)$ and the penalty parameter sequence $\{\l^i\}$. For each parametrization, after running 100 trials, we record the average time used by each algorithm using the \verb#proc.time()# function in the software R~\citep{R}.

\subsection{Implementation of the algorithms} Existing code for the various methods is implemented across different environments (such as C and R). For a fair comparison, therefore, we re-coded the methods in R~\citep{R} using the pseudo-code given in the papers, and implemented the SLS algorithm in R as well.

Code for the method in~\cite{meier2008group} is available via the \verb#grplasso# package in R. For an unbiased comparison with the SLS algorithm, we took our existing code for SLS, and replaced the SLS group update step with the the part of their code pertaining to the group update step. (Except for this update step, the structure of the two algorithms is identical, since both are group-wise descent algorithms).

We coded the algorithms to run on decreasing sequences of penalty values $\{\l^i\}$ for the SLS and the~\cite{meier2008group} algorithms, or increasing sequences of bound values $\{M^i\}$ for the~\cite{kim2006blockwise} and~\cite{roth2008group} algorithms. In each algorithm, convergence at each penalty or bound value is determined by the stopping criterion
$$\|\b^{(t)}-\b^{(t-1)}\|_{\infty}\leq 10^{-8}\;\;.$$

\subsection{Simulated data} We generate the data as follows. In each simulation, we have $n=50$ samples. The number of groups, $K$, ranges in the set $\{10,20,40,80\}$, but the number of groups in the true smallest model is always $2$. Each group has $10$ covariates. The true coefficient vector $\b_0$ is given by
$$(\b_0)_k=\left\{\begin{array}{ll}\mathbf{1}_{10},&k=1,2,\\\mathbf{0}_{10},&k>2\;\;,\\\end{array}\right.$$
where $\mathbf{1}_m$ and $\mathbf{0}_m$ are the vectors in $\mathbb{R}^m$ with all entries equal to $1$ or $0$, respectively.

Each row of $X$ is sampled independently from a $N(\mathbf{0}_n,\Sigma)$ distribution, where $\Sigma$ is determined by two parameters: within-group correlation, $a$, and between-group similarity, $b$, as follows. For $a,b\in[0,1]$, we define $\Sigma$ group-wise as $\Sigma=(\Sigma_{k_1,k_2})_{1\leq k_1,k_2\leq K}$, with:
$$\Sigma_{k_1,k_2}=\left(\begin{array}{cccc}1&a&\dots&a\\a&1&\dots&a\\\dots&\dots&\dots&\dots\\a&a&\dots&1\\   \end{array}\right)\times\left\{\begin{array}{ll}1,&k_1=k_2\\b,&k_1\neq k_2\\\end{array}\right.\;\;.$$

In our experiments, we simulate nine different correlation structures, by pairing\linebreak $a\in\{.2,.5,.8\}$ (low, medium, or high within-group correlation) with $b\in\{.2,.5,.8\}$ (low, medium, or high between-group similarity). We then generate $y\sim N(X\b_0,c^2I_n)$, with $c^2$ defined as
$$c^2=0.01\beta_0^T\Sigma\beta_0=0.01 \cdot Var(x^T\beta_0),$$
where $x\sim N(0,\Sigma)$ represents a single draw of a row of $X$, in order to produce a high signal-to-noise ratio.

\subsection{Penalties and bounds} In practice, it is often useful to compute a `solution path' over a set of values of the penalty parameter $\l$. In fact, \cite{roth2008group} observe that, for their method, given a bound $M$, it is actually often faster to compute a solution path along an increasing sequence of bounds $M^1<M^2<\dots<M^N=M$, than to directly compute the solution for the bound $M$. This sequence of increasing bounds corresponds to a decreasing sequence of penalty parameters, $\l^1>\l^2>\dots>\l^N$. Meier et al's \verb#grplasso# package is also implemented to find solutions along such a solution path, meaning that the solutions for penalty values $\{\l^i\}$ are computed sequentially, using the final solution $\hat{\b}^{\l^i}$ for penalty parameter $\l^i$ as an initial estimate for computing the solution $\hat{\b}^{\l^{i+1}}$ (and using $\mathbf{0}_p$ a an initial estimate for computing $\hat{\b}^{\l^1}$). We use this sequential stucture (with decreasing penalties $\{\l^i\}$ or or increasing bounds $\{M^i\}$, as appropriate) in our implementation of each method.

In each simulation, to choose a sequence of penalty parameters, we first compute $\l_{max}$, defined as
$$\l_{max}=\sup\{\l\geq 0 \ : \ \hat{\b}^{\l}\neq 0\}=\max_k\|X_k^Ty\|_2\;\;,$$
where $\hat{\b}^{\l}$ is any solution to the group lasso problem with penalty parameter $\l$. (The last equality follows from the subdifferential condition). We then choose the sequence $\{\l^i=\l_{max}\times 2^{-i}\}_{1\leq i\leq 5}$.

For each choice of $(X,y)$ and for a value $\l^i$ of the penalty parameter, we find the bound $M^i$ that produces the same solution set by computing $\hat{\b}^{\l^i}$ (via the SLS algorithm, for example) and defining $M^i=\sum_{k=1}^K\|\hat{\b}^{\l^i}_k\|_2$; the solution of~(\ref{eqn:GL_bounded}) with $M=M^i$ is then equal to the solution of~(\ref{eqn:GL}) with $\l=\l^i$. This allows us to compare the ~\cite{kim2006blockwise} and~\cite{roth2008group} algorithms, which use bounds on the sum of group norms, to the SLS and~\cite{meier2008group} algorithms, which use penalties on the group norms.

\subsection{Results}

\begin{figure}[t]
	\centering
		\includegraphics[width=5cm]{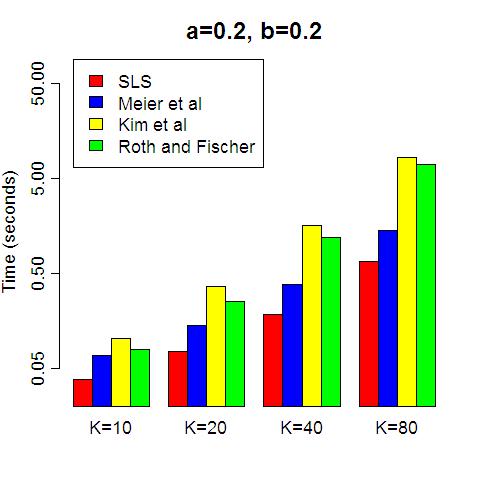}
		\includegraphics[width=5cm]{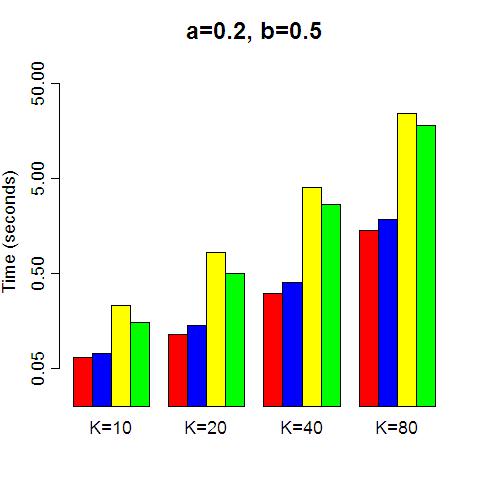}
		\includegraphics[width=5cm]{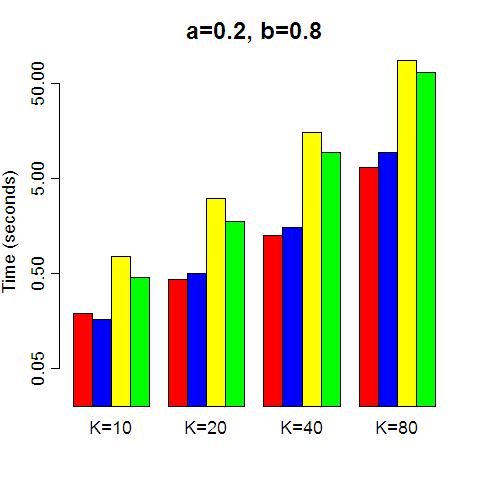}
		\includegraphics[width=5cm]{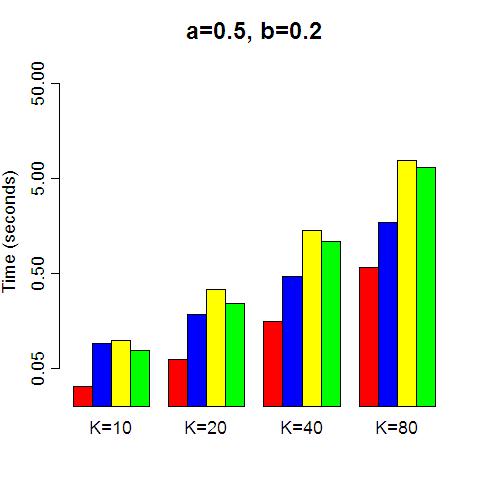}
		\includegraphics[width=5cm]{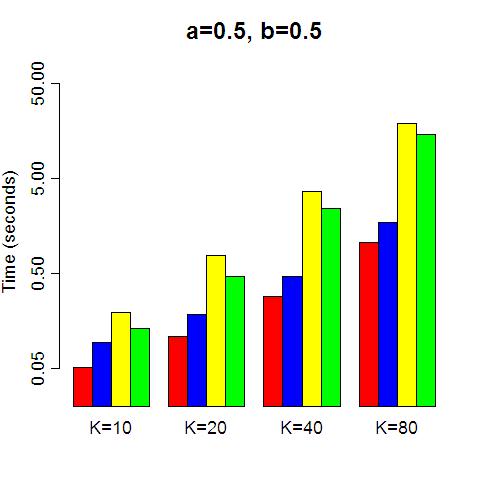}
		\includegraphics[width=5cm]{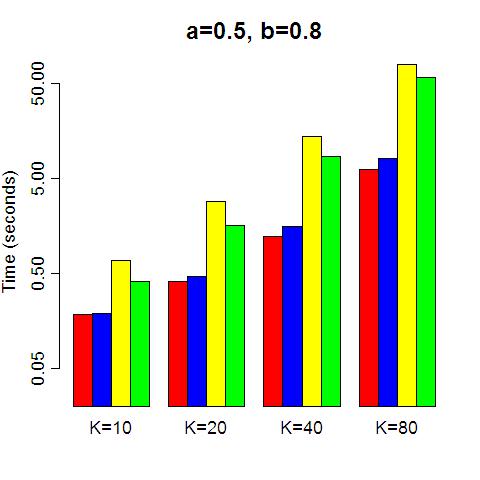}
		\includegraphics[width=5cm]{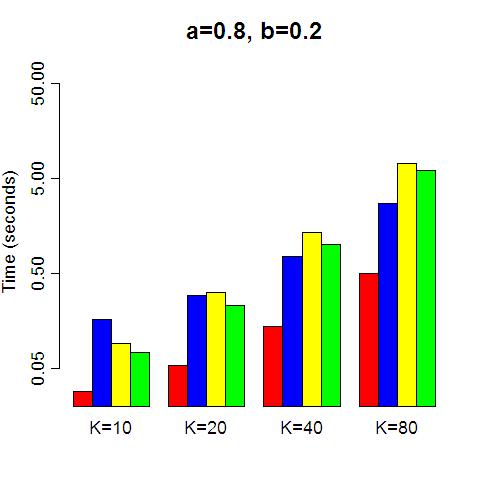}
		\includegraphics[width=5cm]{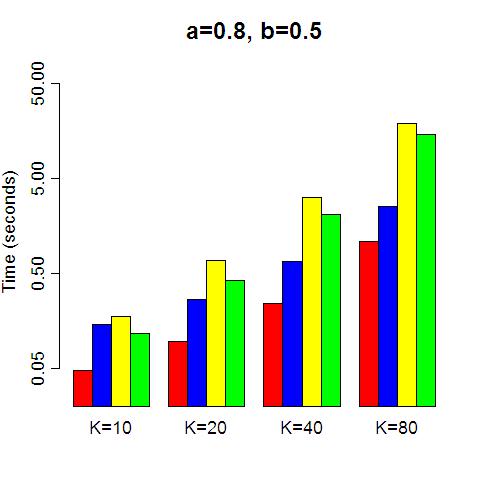}
		\includegraphics[width=5cm]{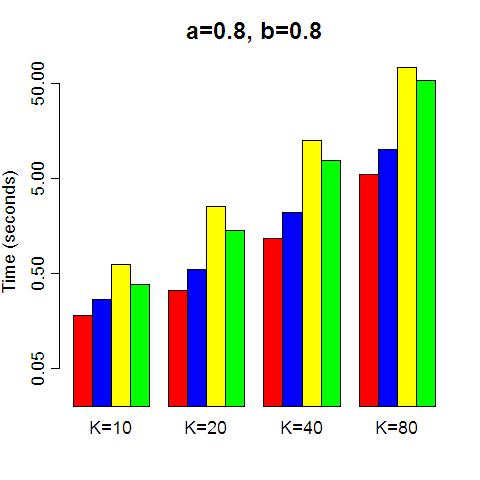}
		\caption{Time until convergence in the group lasso with 10, 20, 40, or 80 total groups ($K$), for the SLS algorithm, the \citet{meier2008group} algorithm, the \citet{kim2006blockwise} algorithm, and the \citet{roth2008group} algorithm. Parameters: $a$=within-group correlation, $b$=between-group similarity. The vertical (time) axis is drawn on a log scale.}
	\label{fig:GL_output_plots}
\end{figure}

Results for our simulations are displayed in Figure~\ref{fig:GL_output_plots} (note that the time axis is drawn on a log scale). Under any choice of parameters $a$, $b$, and $K$, the SLS algorithm converges faster than any of the other four methods considered, with one exception ($a=.2,b=.8,K=10$) when it is slightly outperformed by \citeauthor{meier2008group}'s~\citeyearpar{meier2008group} algorithm. Considering the results as a whole, the most comparable method to the SLS, in terms of performance, is \citeauthor{meier2008group}'s~\citeyearpar{meier2008group} algorithm, which performs almost as fast as the SLS algorithm in some simulations. The SLS algorithm's improvement in speed relative to \citeauthor{meier2008group}'s~\citeyearpar{meier2008group} algorithm is strongest for higher values of $a$ and lower values of $b$. This is intuitive, since a high value of within-group correlation $a$ means that gradient approximations to the group optimization will tend to not be very accurate, and therefore the optimization step of the SLS algorithm is likely to improve time considerably. On the other hand, a high value of between-group similarity $b$ means that many groups will be included at some stage of the algorithm, and so the SLS algorithm will have many spectral decompositions to perform. The remaining two methods are consistently slower than the SLS algorithm in the settings simulated here, and, depending on the setting, slower or comparable to the~\cite{meier2008group} algorithm.

Overall, the efficient structure of the SLS algorithm is clearly evident in its faster computation time relative to the other algorithms, when the number of groups is moderate as in these simulations. (Adapting the algorithm to be efficient in very high-dimensional settings is discussed in Section~\ref{section:Discussion}).

\section{The SSLS algorithm for the sparse group lasso}
\label{section:SSLS}

In recent work, \citet{wu2008coordinate} and~\citet{friedman2010note} discuss the question of within-group sparsity. The original group lasso has the property that, in general settings, with probability $1$, each group of coefficients will be either entirely zero or entirely nonzero in the optimal solution. While this is natural in some settings, there are many settings in which allowing for within-group sparsity would be more plausible, and may help to recover a signal more accurately. The proposed penalized likelihood function \citep{wu2008coordinate,friedman2010note} is given by
$$L_1(\b)=\tfrac{1}{2}\|y-X\b\|^2_2+\l_1\sum_{k=1}^K\|\b_k\|_2+\l_2\|\b\|_1\;\;.$$

\citet{friedman2010note} show that the subdifferential of $L_1$ at $\b$ is separable over the groups, and that the subdifferential of $L_1$ with respect to the $k^{\mathrm{th}}$ group is given by
$$\partial_{\b_k}L_1(\b)=-X_k^T(y-X\b)+\l_1 s_k+\l_2 t_k\;\;,$$
where $s_k=\b_k/\|\b_k\|_2$ whenever $\b_k\neq 0$ and may be any vector of up to unit norm if $\b_k=0$, and $(t_k)_j=\sign((\b_k)_j)$ whenever $(\b_k)_j\neq 0$ and may be any number in $[-1,1]$ if $(\b_k)_j=0$.
A coefficient vector $\hat{\b}$ is therefore a minimizer of $L_1$ if and only if each group's subdifferential contains the zero vector:
$$\hat{\b}\in\arg\min_{\b}L_1(\b) \ \Leftrightarrow \ 0\in\partial_{\b_k}L_1(\hat{\b}) \ \forall k\;\;.$$

We now describe an adaptation of the SLS algorithm, which can solve the sparse group lasso problem effectively for small group sizes. We first explain the intuition behind the algorithm. When updating a single group, the relevant subproblem consists of minimizing

\begin{equation}\label{eqn:subproblem_SGL}Q_1(\a)=\tfrac{1}{2}\|b-A\a\|^2_2+\l_1\|\a\|_2+\l_2\|\a\|_1\;\;,\end{equation}
where $b\in\R^n$, $A\in\R^{n\times q}$, $\l_1,\l_2>0$, and $\a$ may take any value in $\R^q$. We denote the optimum by $\hat{\a}$.

As observed in~\citet{friedman2010note},
$$\hat{\a}=0 \ \Leftrightarrow \ \|\{A^Tb\}_{\l_2}\|_2\leq \l_1\;\;,$$
where $\{\cdot\}$ denotes the soft threshholding operation, defined for a real value $x\in\R$ by $\{x\}_{\l_2}=\sign(x)\cdot\max\{|x|-\l_2,0\}$, and defined on a vector by applying the operation element-wise.
 
When $\a=0$ is not optimal, the subgradient condition for optimality is therefore given by
$$0=-A^Tb+A^TA\a+\l_1 \frac{\a}{\|\a\|_2}+\l_2 t\;\;,$$
where $t_j=\sign(\a_j)$ if $\a_j\neq 0$, and may be any number in $[-1,1]$ if $\a_j=0$. Next, observe that, if $\sign(\hat{\a})$ is known, then we may solve for $\hat{\a}$ via the same strategy as in the SLS algorithm. Specifically, if $\sign(\hat{\a})=\hat{\s}$ for a known $\hat{\s}\in\{-1,0,1\}^q$, then defining $J=\{j \ : \ \hat{\s}_j\neq 0\}$, we know that $\hat{\a}_{J^c}=0$, and that $\hat{\a}_J$ must satisfy
$$0=-A_J^Tb+A_J^T A_J\hat{\a}_J+\l_1\frac{\hat{\a}_J}{\|\hat{\a}_J\|_2}+\l_2\hat{\s}_J\;\;,$$
where $A_J$ is the $n\times|J|$ matrix consisting of the columns of $A$ with indices in $J$. Since $\hat{\s}$ is assumed to be known, we may apply Theorem~\ref{thm:GL} to solve for $\hat{\a}_J$.

In practice, the optimal sign vector $\hat{\s}$ is not known. However, we may cycle through all sign vectors $\s\in\{-1,0,1\}^q$, attempt to solve for $\hat{\a}$ under choice for $\s$, and check for optimality. This intuition is formalized in Theorem~\ref{thm:SGL} below.

\begin{theorem}\label{thm:SGL}
Define
$$Q_1(\a)=\tfrac{1}{2}\|b-A\a\|_2^2+\l_1\|\a\|_2+\l_2\|\a\|_1\;\;,$$
where $b\in\R^n$, $A\in\R^{n\times q}$, $\l_1,\l_2>0$, and $\a$ may take any value in $\R^q$. Then:
\begin{itemize}
\item[i.] Suppose $\|\{A^Tb\}_{\l_2}\|_2\leq \l_1$. Then $\alpha=0$ is the unique minimizer of $Q_1$.
\item[ii.] Suppose $\|\{A^Tb\}_{\l_2}\|_2>\l_1$. For any vector of signs $\s\in\{-1,0,1\}^p$, write  $J=\{j:\s_j\neq0\}$, and let $A_J^TA_J=U^T_JD_JU_J$ be the spectral decomposition, with $D_J=\diag\{d^J_1,\dots,d^J_{|J|}\}$. Define also $v_{\s}=A_J^Tb-\l_2\s_J$ and
$$f_{\s}(r)=\sum_{j=1}^{|J|} \frac{(v_{\s})_j^2}{(d^J_jr+\l_1)^2}\;\;.$$
Define $\hat{\s}\in\{-1,0,1\}^q$ to be the (unknown) vector of signs of the true optimal solution $\hat{\a}=\arg\min_{\a}Q_1(\a)$. Let $\s$ be any vector of signs in $\{-1,0,1\}^q$. Then:
\begin{itemize}
\item[a.] If $\s=\hat{\s}$, there will be exactly one $r$ satisfying $f_{\s}(r)=1$. Furthermore, if we define
$$\alpha_J(r)=U_J^T(D_J+r^{-1}\lambda_1 I_{|J|})^{-1}v_{\s} \ \mathrm{and} \ \a(r)_j=\left\{\begin{array}{ll}(\a_J(r))_j, &j\in J\\0,&j\not\in J\\\end{array}\right.\;\;.$$
then the following feasibility conditions will be satisfied:
$$\sign(\a(r))=\s \ \mathrm{and} \ \forall \ j\not\in J, \ |\{A_j^T(b-A\a)\}_{\l_2}|\leq\l_1\;\;.$$
Moreover, $\hat{\a}=\a(r)$ is the unique minimizer of $Q_1$.
\item[b.] If instead $\s\neq\hat{\s}$, then either $f_{\s}(r)=1$ will have no solutions, or it will have one solution $r$ with $\a(r)$ failing the feasibility conditions.
\end{itemize}
\end{itemize}
\end{theorem}

We are now ready to define the `Signed Single Line Search' algorithm for the sparse group lasso; see the pseudocode in Algorithm~\ref{alg:SSLS}. We note that, at the step updating group $k$, this algorithm could potentially cycle through as many as $3^{p_k}$ sign vectors before finding the optimal group coefficient vector. Therefore we might expect this algorithm to be, at worst, up to $\left(3^{\max_k\{p_k\}}\right)$ times slower than the SLS algorithm. However, cycling through the possible sign vectors may be done in an order that is better than random, lowering the expected number of sign vectors which needs to be tested at each step.

By Theorem~\ref{thm:SGL}, we know that at each step, the algorithm finds the optimal value for $\b_k$ (conditional on the other groups' coefficient estimates at that time). We therefore have the immediate corollary:

\begin{corollary}\label{cor:decrease_SGL}
Let $\b^{(t)}$ be the coefficient estimate after $t$ iterations of the SSLS algorithm for $t=0,1,2,\dots$. Then for all $t$,
$$L_1(\b^{(t+1)})\leq L_1(\b^{(t)})\;\;.$$
That is, each iteration of the algorithm does not increase the objective function.
\end{corollary}

Finally, as with the SLS algorithm, we state convergence and accuracy results. The convergence result again follows directly from Proposition 5.1 of~\citet{tseng2001convergence}. The proof of the accuracy theorem is very similar to that of Theorem~\ref{thm:accuracy_GL}, and we omit it in this paper.

\begin{theorem}\label{thm:convergence_SGL}
Let  $\b^{(t)}$ be the coefficient vector after the $t^{th}$ iteration of Algorithm~\ref{alg:SSLS}.  Then $L_1(\b^{(t)})\rightarrow\min_{\b}L_1(\b)$.
\end{theorem}

\begin{theorem}\label{thm:accuracy_SGL}
Take any $\b^*\in\R^p$, and any $w^*\in \partial L_1(\b^*)$. Let $\hat{y}$ be the unique optimal vector of fitted values. Then
$$\|X\b^*-\hat{y}\|^2_2\leq 2(w^*)^T\b^*+\mathbf{O}(\|w^*\|_2)\;\;.$$
\end{theorem}

Lacking competing methods to compare to, we do not report on numerical experiments with the SSLS algorithm. However, we remark that an R implementation solved sparse group lasso problems with $40$ groups of size $5$ (and with a single choice of penalty parameters $(\l_1,\l_2)$ which produced appropriate sparsity patterns) in a few seconds.

\begin{algorithm}[t]
\caption{Signed Single Line Search (SSLS) algorithm for the sparse group lasso}
\label{alg:SSLS}
\begin{algorithmic}
\STATE \textbf{Input:} $y\in\R^n$, $X_1\in\R^{n\times p_1},\dots, X_K\in\R^{n\times p_K}$, $\l_1,\l_2>0$.
\STATE \textbf{Output:} $\b$ minimizing $L_1(\b)=\tfrac{1}{2}\|y-X\b\|_2^2+\l_1\sum_{k=1}^K\|\b_k\|_2+\l_2\|\b\|_1$.\\
\STATE \textbf{Initialize:} $\b\Leftarrow \b^{(0)}$.\\
\REPEAT
\FOR{$k=1,2,\dots,K$} 
\STATE $R_k\Leftarrow y-\sum_{l\neq k}X_l\b_l$.
\IF{$\|\{X_k^TR_k\}_{\l_2}\|_2\leq \l_1$}
\STATE $\b_k\Leftarrow 0$.
\ELSE
\REPEAT
\STATE Choose some sign vector $\s\in\{-1,0,1\}^{p_k}$.
\STATE Solve the optimization problem associated with $\s$ as in Procedure~\ref{alg:SSLS_subroutine}.
\UNTIL a feasible solution has been found
\ENDIF
\ENDFOR
\UNTIL{some convergence criterion is met.}
\end{algorithmic}
\end{algorithm}

\floatname{algorithm}{Procedure}
\begin{algorithm}[t]
\caption{SSLS subroutine}
\label{alg:SSLS_subroutine}
\begin{algorithmic}
\STATE $J\Leftarrow\{j \ : \ \s_j\neq 0\}$.
\STATE Compute the spectral decomposition $(X_k)_J^T(X_k)_J=U_{k;J}^TD_{k;J}U_{k;J}$ if not previously computed, and write $D_{k;J}=\diag\{d^{k;J}_1,\dots,d^{k;J}_{p_k}\}$.
\STATE $v_{k;\s}\Leftarrow U_{k;J}\left((X_k)_J^TR_k-\l_2\s_J\right)$.
\IF{There exists an $r$ satisfying $f(r)=\sum_{j=1}^{|J|}\frac{(v_{k;\s})_j^2}{(d^{k;J}_jr+\l_1)^2}=1$}
\STATE $\a\Leftarrow U_{k;J}^T(D_{k;J}+r^{-1}\l_1I_{|J|})^{-1}v_{k;\s}$.
\IF{$\sign(\a)=\s_J$, and for all $j\not\in J$, $|\{(X_k)_j^T(R_k-(X_k)_J\a)\}_{\l_2}|\leq \l_1$}
\STATE $(\b_k)_j\Leftarrow \a_j$ for all $j\in J$.
\STATE $(\b_k)_j\Leftarrow 0$ for all $j\not\in J$.
\STATE This solution is feasible.
\ELSE
\STATE No feasible solution exists.
\ENDIF
\ELSE
\STATE No feasible solution exists.
\ENDIF
\end{algorithmic}
\end{algorithm}

\section{Discussion}
\label{section:Discussion}

For the group lasso in the linear regression setting, the SLS algorithm offers a fast and exact group-wise update step, which, in our simulations, performs very well on moderately-sized problems as compared to existing methods. One immediate extension of the SLS algorithm would be to make use of the `active set' construction developed by~\citet{roth2008group}, the framework of which may be combined with any algorithm that solves the group lasso problem. Their work shows that adding this `active set' construction to \citeauthor{kim2006blockwise}'s~\citeyearpar{kim2006blockwise} global descent algorithm may speed up computation considerably. Combining the SLS algorithm with \citeauthor{roth2008group}'s~\citeyearpar{roth2008group} `active set' construction is therefore likely to improve computation speed on very large (and very sparse) group lasso problems. Furthermore, while this paper's focus is on linear regression, our methods may be extended to other likelihood functions via second-order approximations, as in the work on the logistic case in~\citet{meier2008group}. However, since any other likelihood function will not be exactly quadratic (as in the case of linear regression), our method will not be able to solve directly for each group's optimum value (fixing the other groups' coefficients), and so it is not clear whether an improvement in speed can be expected in non-linear regression.

In the case of the sparse group lasso, there are many possibilities for developing a more efficient algorithm based on the same principles as the SLS algorithm for group lasso. The strategy of exhaustive search through sign configurations is, of course, impractical for even a moderately large group size. One alternative approach is to reduce to single-coordinate descent rather than group-wise descent (in order to avoid the issue of sign configurations). However, this is potentially problematic, because any coordinate descent approach to either a group lasso or sparse group lasso problem has the drawback of occasionally converging to a non-optimal solution. Specifically, the coefficients within some group $k$ may become `trapped' at zero, even when $\b_k=0$ is not optimal, due to the structure of the $2$-norm penalty. We illustrate this with an example; the example is phrased in the group lasso setting, but may easily be adapted to show that the same problem may occur in the sparse group lasso setting.

\begin{example}\label{ex:zerotrap} {\bf(The `zero trap')}. Consider a group lasso problem with a single group consisting of two covariates. Define
$$y=\left(\begin{array}{c}1\\1\\\end{array}\right), \ X=\left(\begin{array}{cc}1&0\\0&1\\\end{array}\right), \ \l=1\;\;.$$
With these data and parameter values, the objective function in~(\ref{eqn:GL}) becomes:
$$L(\b)=\tfrac{1}{2}(1-\b_1)^2+\tfrac{1}{2}(1-\b_2)^2+\|\b\|_2\;\;.$$
If we fix $\b_2=0$ and optimize over $\b_1$, we obtain $\b_1=0$. If we then update $\b_2$, we obtain $\b_2=0$. Therefore, coordinate descent with a starting value of $\b_1=\b_2=0$ will never leave this value. However, the value of $\b$ which minimizes $L$ is given by $\b_1=\b_2=1-\frac{\sqrt{2}}{2}$.
\end{example}

We conclude that, in any situation where some groups might have a signal which is weak on any individual coefficient but significant in total, coordinate descent methods of optimization may not be reliable, and thus requires extra care to allow us to circumvent the problem of sign configurations. 

In the SSLS algorithm, any given update of the $k^{\mathrm{th}}$ group may need to test up to $3^{p_k}$ sign configurations. However, when the algorithm has neared the true solution, we might expect `sign stabilization'; that is, the optimal sign vector at iteration $t$ may be unchanged at iteration $(t+1)$. This suggests that attempting a signed single-line-search update for each group may be very efficient after a certain point. For early iterations, when many groups are not yet `sign stabilized', other methods (such as gradient-based methods) could be considered. The potential efficiency of this kind of algorithm lies in the fact that whenever a group $k$ has achieved sign stabilization, the algorithm could optimize the entire group of coefficients at once rather than pursuing any less efficient update strategy. We plan to develop this strategy in future work in order to create an algorithm that can solve the sparse group lasso problem with moderate or large group size.

\section{Appendix}\label{section:Appendix}

In this section we prove all theoretical results stated in the paper. In order for this appendix to be self-contained we restate some of the theorems.\\

\noindent
{\bf Theorem \ref{thm:solution_set_GL}}
{\it Let $\hat{B}$ be the set of minimizers of the penalized likelihood $L$ for a group lasso problem~(\ref{eqn:GL}) or sparse group lasso problem~(\ref{eqn:SGL}). Then there exists a unique minimal set of groups $\mathcal{K}\subset\{1,\dots,K\}$, and unique unit vectors $v_k\in\R^{p_k}$ for each $k\in\mathcal{K}$, such that
$$\hat{\b}\in\hat{B} \ \Rightarrow \ \hat{\b}_k\posprop v_k \ \forall k\in\mathcal{K}, \ \hat{\b}_k=0 \ \forall k\not\in\mathcal{K}\;\;,$$
where we define $a\posprop b$ to mean that $a=c\cdot b$ for some nonnegative scalar $c$.
}

\begin{proof}
(This proof addresses the sparse group lasso case; the group lasso case can be obtained by setting $\l_1=\l$ and $\l_2=0$ in the sparse group lasso problem).

We first make an observation about the convexity of $L(\b)$. Take any $\b^1,\b^2\in\R^p$. If $\b^1_k,\b^2_k\neq 0$, and there does not exist a positive $c$ with $\b^1_k=c\b^2_k$, then $\|\b^1_k+\b^2_k\|_2<\|\b^1_k\|_2+\|\b^2_k\|_2$. Therefore, since all other terms in $L$ are convex in $\b$, we know that $L(\tfrac{1}{2}(\b^1+\b^2))<\tfrac{1}{2}(L(\b^1)+L(\b^2))$. This implies that for any $\b^1,\b^2\in\hat{B}$, if $\b^1_k,\b^2_k\neq 0$, then there must exist a positive $c$ with $\b^1_k=c\b^2_k$. Now define $\mathcal{K}$ as:
$$\mathcal{K}=\{k \ : \ \exists\hat{\b}\in\hat{B}, \ \b_k\neq 0\}\;\;.$$
Then, for each $k\in\mathcal{K}$, we can find a unique unit vector $v_k$, such that for any $\b\in\hat{B}$, $\b_k=cv_k$ for some $c\geq 0$. The uniqueness and minimality of $\mathcal{K}$ are clear from its definition.
\end{proof}

\noindent
{\bf Theorem \ref{thm:GL}}
{\it Define
$$Q(\a)=\tfrac{1}{2}\|b-A\a\|_2^2+\l\|\a\|_2\;\;,$$
where $b\in\R^n$, $A\in\R^{n\times q}$, $\l>0$, and $\a$ may take any value in $\R^q$.
Let $A^TA=U^TDU$ be the spectral decomposition, with $D=\diag\{d_1,d_2,\dots,d_q\}$. Define $v=UA^Tb$. Then:
\begin{itemize}
\item[i.] If $\|v\|_2\leq \lambda$ then $\alpha=0$ is the unique minimizer of $Q$.
\item[ii.] If $\|v\|_2>\lambda$, then there is a unique $r\in\mathbb{R}_+$ satisfying
\begin{equation}\label{eqn:solve_for_r}f(r)=\sum_{j=1}^q \frac{v_j^2}{(d_jr+\lambda)^2}=1\;\;.\end{equation}
Furthermore, if we define
\begin{equation}\label{eqn:solve_for_alpha}\alpha(r)=U^T(D+r^{-1}\lambda I_p)^{-1}v\;\;,\end{equation}
then $\alpha=\alpha(r)$ is the unique minimizer of $Q$.
\end{itemize}
}

\begin{proof}
The minimizer of $Q$ is unique due to strict convexity of $Q$. \citet{friedman2010note} discuss the subgradient of $Q(\a)$, and the implication that $\a=0$ minimizes $Q$ if and only if $\|v\|_2\leq \l$; this covers the first case.
Assume that the second case holds; that is, $\|v\|_2>\lambda$. By~\citet{friedman2010note}, $\alpha$ is the unique minimizer if and only if it satisfies the subgradient equation
\begin{equation}\label{eqn:subgradient}A^TA\alpha+\lambda \frac{\alpha}{\|\alpha\|_2}=A^Tb\;\;.\end{equation}
Since $\|v\|_2>\lambda$, referring to~(\ref{eqn:solve_for_r}), we see that $f(0)>1$ and $\lim_{r\rightarrow\infty}f(r)=0$. Since $f$ is strictly decreasing in $r$, then there is a unique $r>0$ with $f(r)=1$. (To check that $\lim_{r\rightarrow\infty}f(r)=0$, we compute the singular value decomposition $A=V^TD^{1/2}U$, where $(D^{1/2})^TD^{1/2}=D$ in the notation of the theorem. Then $v=UA^Tb=(D^{1/2})^TVb$, and so for any $j$ with $d_j=0$, $v_j=0$ also. Therefore, $f(r)$ vanishes as $r\rightarrow\infty$).
Let $\alpha=\alpha(r)$. By ~(\ref{eqn:solve_for_r}) and~(\ref{eqn:solve_for_alpha}),
$$r^2=r^2\left(\sum_{j=1}^q \frac{v_j^2}{(d_jr+\lambda)^2}\right)=\sum_{j=1}^q \frac{v_j^2}{(d_j+r^{-1}\lambda)^2}=v^T(D+r^{-1}\lambda I_p)^{-2}v=\|\alpha\|_2^2\;\;.$$
Therefore, we can rewrite~(\ref{eqn:solve_for_alpha}) as
$$\left(D+\frac{\lambda}{\|\alpha\|_2}I_p\right)U\alpha=UA^Tb\;\;.$$
Hence,
$$U^T\left(D+\frac{\lambda}{\|\alpha\|_2}I_p\right)U\alpha=\left(U^TDU+\frac{\lambda}{\|\alpha\|_2}U^TU\right)\alpha=A^TA\alpha+\lambda\frac{\alpha}{\|\alpha\|_2}=A^Tb\;\;.$$
This proves that $\a$ satisfies~(\ref{eqn:subgradient}) and is thus the unique minimizer of $Q$.
\end{proof}

\noindent
{\bf Theorem \ref{thm:accuracy_GL}}
{\it 
Take any $\b^*\in\R^p$, and any $w^*\in \partial L(\b^*)$. Let $\hat{y}$ be the unique optimal vector of fitted values. Then
$$\|X\b^*-\hat{y}\|^2_2\leq 2(w^*)^T\b^*+\mathbf{O}(\|w^*\|_2)\;\;.$$
More precisely, the error in the estimate of the optimal fitted values is bounded as follows, where $\hat{\b}$ is any vector in $\hat{B}$:
$$\|X\b^*-\hat{y}\|^2_2\leq 2(w^*)^T\b^*+2\|w^*\|_2\|\hat{\b}\|_2\;\;.$$
By bounding $\|\hat{\b}\|_2$, we may further obtain the following two bounds (here $\b_{LSE}$ denotes any unpenalized least-squares estimate minimizing $\|y-X\b_{LSE}\|_2^2$):
$$\|X\b^*-\hat{y}\|^2_2\leq 2(w^*)^T\b^*+2\|w^*\|_2\times \l^{-1}\left(L(\b^*)-\tfrac{1}{2}\|P_{X}^{\perp}y\|^2_2\right)\;\;.$$
$$\|X\b^*-\hat{y}\|^2_2\leq 2(w^*)^T\b^*+2\|w^*\|_2\times \sum_{k=1}^K \|(\b_{LSE})_k\|_2\;\;.$$
}
\begin{proof}
First, take any any $\b$, any $\d\in\R^p$, and any subgradient $w\in\partial L(\b)$. Observe that
$$\tfrac{1}{2}\|y-X(\b+\d)\|^2_2=\tfrac{1}{2}\|y-X\b\|^2_2-(y-X\b)^T\d+\tfrac{1}{2}\|X\d\|^2_2\;\;,$$
and from the gradient of the likelihood,
$$w\in\partial L(\b) \ \Rightarrow \ \left(w+(y-X\b)\right)\in\partial \left(\l\sum_{k=1}^K\|\b_k\|_2\right)\;\;.$$
By the definition of the subdifferential,
$$\l\sum_{k=1}^K\|(\b+\d)_k\|_2\geq \l\sum_{k=1}^K\|\b\|_2+\left(w+(y-X\b)\right)^T\d\;\;.$$
Therefore,
\begin{eqnarray*}
L(\b+\d)&=&\tfrac{1}{2}\|y-X(\b+\d)\|^2_2+\l\sum_{k=1}^K\|(\b+\d)_k\|_2\\
&\geq&\tfrac{1}{2}\|y-X\b\|^2_2-(y-X\b)^T\d+\tfrac{1}{2}\|X\d\|^2_2+\l\sum_{k=1}^K\|\b\|_2+\left(w+(y-X\b)\right)^T\d\\
&=&L(\b)+w^T\d+\tfrac{1}{2}\|X\d\|^2_2\;\;.\\
\end{eqnarray*}
Now take $w^*\in\partial L(\b^*)$ and $\hat{\b}\in\arg\min_{\b}L(\b)$. From above,
$$L(\hat{\b})\geq L(\b^*)+(w^*)^T(\hat{\b}-\b^*)+\tfrac{1}{2}\|X(\hat{\b}-\b^*)\|^2_2\;\;.$$
Also, by optimality of $\hat{\b}$, $L(\hat{\b})\leq L(\b^*)$. Therefore,
$$\tfrac{1}{2}\|X\b^*-\hat{y}\|^2_2\leq (w^*)^T(\b^*-\hat{\b})\leq (w^*)^T\b^*+\|w^*\|_2\|\hat{\b}\|_2\;\;.$$
We next observe that, for any $\b\in\R^p$,
\begin{equation}\label{eqn:beta_hat_norm}L(\b)\geq L(\hat{\b})\geq \tfrac{1}{2}\|P_X^{\perp}y\|^2_2+\l\sum_{k=1}^K\|\hat{\b}_k\|_2\;\;.\end{equation}
Choosing $\b=\b^*$ and applying~(\ref{eqn:beta_hat_norm}), we obtain
$$\|\b\|_2\leq\sum_{k=1}^K  \|\hat{\b}_k\|_2\leq \l^{-1}\left(L(\b^*)-\tfrac{1}{2}\|P_X^{\perp}y\|^2_2\right)\;\;,$$
which yields the next-to-last bound in the theorem. Choosing instead $\b=\b_{LSE}$ and again applying~(\ref{eqn:beta_hat_norm}), using the fact that $\|y-X\b_{LSE}\|^2_2=\|P_X^{\perp}y\|^2_2$, we obtain
$$\|\b\|_2\leq\sum_{k=1}^K  \|\hat{\b}_k\|_2\leq \l^{-1}\left(L(\b_{LSE})-\tfrac{1}{2}\|P_X^{\perp}y\|^2_2\right)=\sum_{k=1}^K\|\b_{LSE}\|_2\;\;,$$
which yields the last bound in the theorem.
\end{proof}

\begin{remark}
If $\b^*=\b^{(t)}$ for some large $t$, then $L(\b^*)$ will potentially be much lower than $L(\b_{LSE})$. Therefore, the next-to-last bound in the statement of the theorem will be advantageous. It might be possible that the bound can be improved to a bound that does not include $\l^{-1}$ term, but we have not been able to prove this.
\end{remark}

\noindent
{\bf Theorem \ref{thm:SGL}}
{\it 
Define
$$Q_1(\a)=\tfrac{1}{2}\|b-A\a\|_2^2+\l_1\|\a\|_2+\l_2\|\a\|_1\;\;,$$
where $b\in\R^n$, $A\in\R^{n\times q}$, $\l_1,\l_2>0$, and $\a$ may take any value in $\R^q$. Then:
\begin{itemize}
\item[i.] Suppose $\|\{A^Tb\}_{\l_2}\|_2\leq \l_1$. Then $\alpha=0$ is the unique minimizer of $Q_1$.
\item[ii.] Suppose $\|\{A^Tb\}_{\l_2}\|_2>\l_1$. For any vector of signs $\s\in\{-1,0,1\}^p$, write  $J=\{j:\s_j\neq0\}$, and let $A_J^TA_J=U^T_JD_JU_J$ be the spectral decomposition, with $D_J=\diag\{d^J_1,\dots,d^J_{|J|}\}$. Define also $v_{\s}=A_J^Tb-\l_2\s_J$ and
\begin{equation}\label{eqn:solve_for_r_sign}f_{\s}(r)=\sum_{j=1}^{|J|} \frac{(v_{\s})_j^2}{(d^J_jr+\l_1)^2}\;\;.\end{equation}
Define $\hat{\s}\in\{-1,0,1\}^q$ to be the (unknown) vector of signs of the true optimal solution $\hat{\a}=\arg\min_{\a}Q_1(\a)$. Let $\s$ be any vector of signs in $\{-1,0,1\}^q$. Then:
\begin{itemize}
\item[a.] If $\s=\hat{\s}$, there will be exactly one $r$ satisfying $f_{\s}(r)=1$. Furthermore, if we define
\begin{equation}\label{eqn:solve_for_alpha_sign}\alpha_J(r)=U_J^T(D_J+r^{-1}\lambda_1 I_{|J|})^{-1}v_{\s} \ \mathrm{and} \ \a(r)_j=\left\{\begin{array}{ll}(\a_J(r))_j, &j\in J\\0,&j\not\in J\\\end{array}\right.\;\;.\end{equation}
then the following feasibility conditions will be satisfied:
\begin{equation}\label{eqn:feasibility_sign}\sign(\a(r))=\s \ \mathrm{and} \ \forall \ j\not\in J, \ |\{A_j^T(b-A\a)\}_{\l_2}|\leq\l_1\;\;.\end{equation}
Moreover, $\hat{\a}=\a(r)$ is the unique minimizer of $Q_1$.
\item[b.] If instead $\s\neq\hat{\s}$, then either $f_{\s}(r)=1$ will have no solutions, or it will have one solution $r$ with $\a(r)$ failing the feasibility conditions.
\end{itemize}
\end{itemize}
}

\begin{proof}
The question of whether $\a=0$ is optimal is directly covered in~\citet{friedman2010note}. Now assume $\a=0$ is not optimal. For $\a\neq 0$, by~\citet{friedman2010note}, we have:
\begin{equation}\label{eqn:subgradient_sign}\partial Q(\a)=-A^Tb+A^TA\a+\l_1\frac{\a}{\|\a\|_2}+\l_2 t,\end{equation}
where $t_j=\sign(\a_j)$ if $\a_j\neq0$, and may equal any number in $[-1,1]$ if $\a_j=0$.

First we examine the true solution $\hat{\a}$, with its sign vector $\hat{\s}$; define $U_J,D_J,v_{\hat{\s}},f_{\hat{\s}}$ as in the statement of the theorem. We see that
\begin{eqnarray*}
0=(\partial Q(\a))_J&=&-A_J^Tb+A_J^TA\a+\l_1\left(\frac{\a}{\|\a\|_2}\right)_J+\l_2t_J\\
&=&-A_J^Tb+A_J^TA_J\a_J+\l_1\frac{\a_J}{\|\a_J\|_2}+\l_2\hat{\s}_J\;\;.\\
\end{eqnarray*}
It follows that
$$\left(A_J^TA_J+\frac{\l_1}{\|\a_J\|_2}I_{|J|}\right)\a_J=A_J^Tb-\l_2\hat{\s}_J\;\;.\\$$

Next, as discussed in the proof of Theorem~\ref{thm:GL}, for any $\s$, $f_{\s}$ is strictly decreasing in $r$ with $\lim_{r\rightarrow\infty} f_{\s}(r)=0$. Therefore, there is at most one $r>0$ with $f_{\s}(r)=1$. 

Next we check that setting $r=\|\hat{\a}_J\|_2=\|\hat{\a}\|_2$ must satisfy $f_{\hat{\s}}(r)=1$. Indeed,
\begin{eqnarray*}f_{\hat{\s}}(r)&=&\sum_{j=1}^{|J|} \frac{(v_{\s})_j^2}{(d^{J}_jr+\l_1)^2}=\frac{1}{r^2}\sum_{j=1}^{|J|} \frac{(v_{\s})_j^2}{(d^{J}_j+\l_1r^{-1})^2}\\
&=&\frac{1}{r^2}\|(D_{J}+r^{-1}\l_1I_{|J|})^{-1}v_{\s}\|_2^2=\frac{1}{r^2}\|\hat{\a}\|^2_2=1\;\;.\\
\end{eqnarray*}
And, by definition, we would then have $\hat{\a}=\a(r)$. Furthermore, $\sign(\hat{\a})=\hat{\s}$ by assumption. Finally, for all $j\not\in J$, since $\hat{\a}_j=0$ and $\hat{\a}$ is the optimal solution, by~\ref{eqn:subgradient_sign},
$$-A_j^T(b-A\hat{\a})=-A_j^Tb+A_j^TA\hat{\a}+\l_1\frac{\hat{\a}_j}{\|\hat{\a}\|_2}\in\l_2\times[-1,1]\;\;.$$
Therefore, feasibility conditions~(\ref{eqn:feasibility_sign}) hold.

Conversely, take any arbitrary sign vector $\s\in\{-1,0,1\}^p$ and define $U_J,D_J,v_{\s},f_{\s}$ as stated above. Suppose some $r\geq0$ satisfies $f_{\s}(r)=1$, and define $\a=\a(r)$; suppose furthermore that the feasibility conditions~(\ref{eqn:feasibility_sign}) hold. Then by~(\ref{eqn:solve_for_r_sign}) and~(\ref{eqn:solve_for_alpha_sign}), proceeding as in Theorem~\ref{thm:GL}, we see that $\|\a\|^2_2=r^2$, and
$$A_J^TA\a+\l_1\frac{\a_J}{\|\a\|_2}=v_{\s}=A_J^Tb-\l_2\s_J=A_J^Tb-\l_2\sign(\a_J)=A_J^Tb-\l_2t_J\;\;.$$
Moreover, for all $j\not\in J$, since $\a_j=0$ and the feasibility conditions~(\ref{eqn:feasibility_sign}) are satisfied, we have that
$$t_j:=\l_1^{-1}\left(-A_j^Tb+A_j^TA\a+\l_1\frac{\a_j}{\|\a\|_2}\right)\in [-1,1]\;\;.$$
Therefore, with this definition of $t$, we see that $\a=\a(r)$ gives a zero subgradient for $Q$ in~(\ref{eqn:subgradient_sign}), therefore $\a$ is the unique minimizer of $Q(\a)$. Therefore, $\a(r)=\hat{\a}$ and $\s=\sign(\a(r))=\sign(\hat{\a})=\hat{\s}$.
\end{proof}

\bibliographystyle{plainnat} 
\bibliography{GLbibliography}

\end{document}